\documentclass[11pt,a4paper]{article}
\usepackage[hyperref]{./acl2019-latex/acl2019}
\usepackage{times}
\usepackage{latexsym}
\usepackage{enumitem}
\usepackage{graphicx}
\usepackage{amsmath}

\usepackage{tabu}
\usepackage{booktabs}

\usepackage{subcaption}
\usepackage{xcolor}
\usepackage{xspace}
\usepackage{multirow}
\usepackage{lipsum}
\usepackage{tikz}

\usepackage{shortcuts}
\usepackage{lstm}

\newcommand{\enotesoff}{\long\gdef\enote##1##2{}}
\newcommand{\enoteson}{\long\gdef\enote##1##2{{\texttt{$<<<$ ##1: ##2 $>>>$\newline}}}}
\enoteson
\enotesoff
\title{The (Non-)Utility of Structural Features in \lstm-based Dependency~Parsers}

\aclfinalcopy % Uncomment this line for the final submission
\def\aclpaperid{1690} %  Enter the acl Paper ID here

\author{Agnieszka Falenska \and Jonas Kuhn \\
      Institut f\"{u}r Maschinelle Sprachverarbeitung \\
      University of Stuttgart \\
      {\tt name.surname@ims.uni-stuttgart.de}}

\date{}

\begin{document}
\maketitle

\begin{abstract}
Classical non-neural dependency parsers put considerable effort on the design of feature functions. Especially, they benefit from information coming from structural features, such as features drawn from neighboring tokens in the dependency tree.
In contrast, their BiLSTM-based successors achieve state-of-the-art performance without explicit information about the structural context. In this paper we aim to answer the question: How much structural context are the BiLSTM representations able to capture implicitly?
We show that features drawn from partial subtrees become redundant when the BiLSTMs are used. 
We provide a deep insight into information flow in transition- and graph-based neural architectures to demonstrate where the implicit information comes from when the parsers make their decisions. Finally, with model ablations 
we demonstrate that the structural context is not only present in the models, but it significantly influences their performance.
\end{abstract}

\begin{figure*}[t]
\centering
\begin{subfigure}[b]{.53\textwidth}
  \centering
  \includegraphics[width=0.97\linewidth]{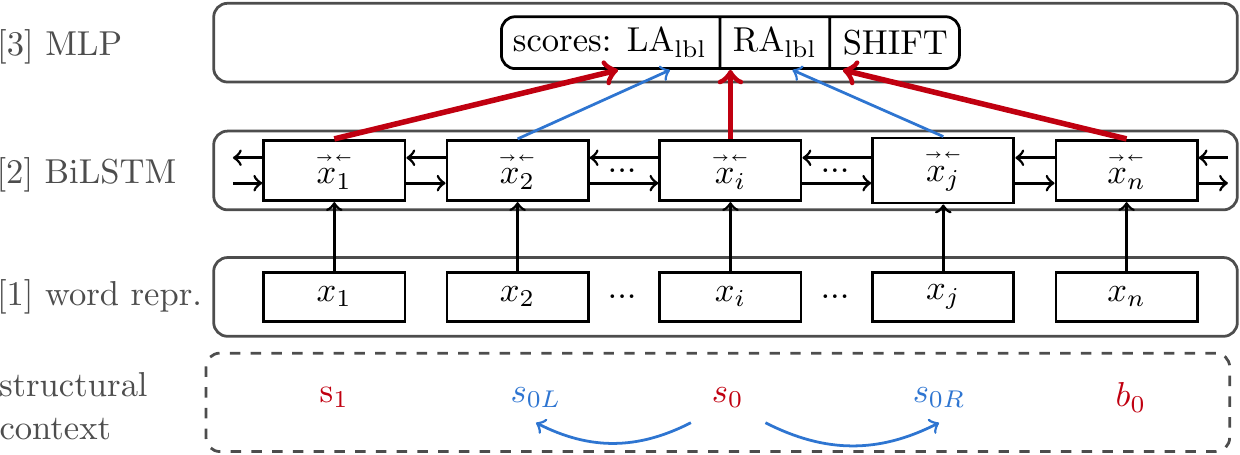}
  \caption{Transition-based parser; scoring transitions for the configuration  $\langle\Sigma, B, A\rangle = \langle x_1 \ldots x_i, x_n, \{ x_i \rightarrow x_2, x_i \rightarrow x_j, \ldots \}\rangle$}
  \label{fig:arch-trans}
\end{subfigure}\hfill
\begin{subfigure}[b]{.445\textwidth}
  \centering
  \includegraphics[width=0.97\linewidth]{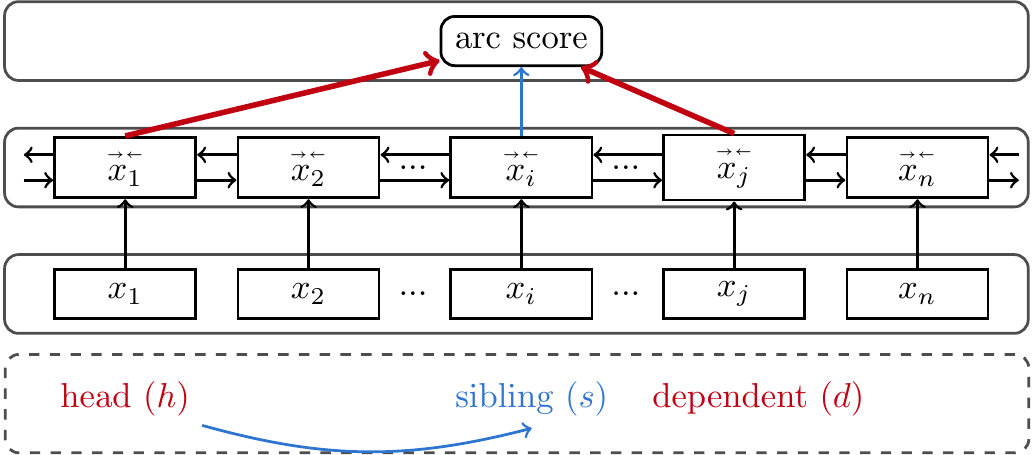}
  \caption{Graph-based parser; \\scoring an arc $x_1 \rightarrow x_j$\\}
  \label{fig:arch-graph}
\end{subfigure}
\caption{Schematic illustration of the \kiparch architecture of \lstm-based neural dependency parsers. Red arrows mark the basic feature sets and blue show how to extend them with features drawn from structural context.}
\label{fig:architecture}
\end{figure*}

\section{Introduction}

%%%
% conventional feature extraction

When designing a conventional non-neural parser substantial effort is required to design a powerful feature extraction function. Such a function \cite[among others]{mcdonald-EtAl:2005:ACL,zhang-nivre:2011:HLT} is constructed so that it captures as much \emph{structural context} as possible. The context allows the parser to make well-informed decisions.\footnote{See Figure~\ref{fig:architecture} for the concept of structural context, details of the architectures will be described in Section~\ref{sec:models}.} It is encoded in features built from partial subtrees and \emph{explicitly} used by the models.
 
%%%
% neural parsers

Recently, \newcite[\kiparch]{kiperwasser-goldberg:2016:TACL-a} showed that the conventional feature extraction functions can be replaced by modeling the left- and right-context of each word with \lstms \cite{hochreiter-schmidhuber:1997:NP,graves-schmidhuber:2005:NN}. Although the proposed models \emph{do not use any conventional structural features} they achieve state-of-the-art performance. The authors suggested that it is because the \lstm encoding is able to estimate the missing information from the given features and did not explore this issue further.

Since the introduction of the \kiparch architecture \lstm-based parsers have become
standard in the field.\footnote{See results from the recent CoNLL 2018 shared task on dependency parsing \cite{zeman-etal:2018:CoNLL} for a comparison of various high-performing dependency parsers.} Yet, it is an open question how much conventional structural context the \lstms
representations actually are able to capture \emph{implicitly}. Small architectures that ignore the structural context are attractive since they come with lower time complexity. But to build such architectures it is important to investigate to what extent the explicit structural information is redundant. 
For example, \kiparch also proposed an extended feature set derived from
structural context, which has subsequently been re-implemented and used by others without questioning its utility.

%%%
% need to understand why it works
% + contributions

Inspired by recent work~\cite{gaddy-etAl:2018:NAACL} on constituency parsing we aim at understanding what type of information is captured by the internal representations of \lstm-based dependency parsers and how it translates into their impressive accuracy. As our starting point we take the \kiparch architecture and extend it with a second-order decoder.\footnote{To the best of our knowledge, this is the first \lstm-based second-order dependency parser. \newcite{gomez-rodriguez-etal-2018-global} incorporate \lstm-based representations into the third-order 1-Endpoint-Crossing parser of \newcite{pitler-2014-crossing}.} We perform systematic analyses on nine languages using two different architectures 
(transition-based and graph-based) across two dimensions: with and without \lstm representations, and with and without features drawn from structural context.

We demonstrate that structural features are useful for neural dependency parsers but they become redundant when \lstms are used (Section~\ref{sec:testresults}). 
It is because the \lstm representations trained together with dependency parsers capture a significant amount of complex syntactic relations
(Section~\ref{sec:representation_analysis}). 
We then carry out an extensive investigation of information flow in the parsing architectures and find that the
implicit structural context is not only present in the \lstm-based parsing models, but also more diverse than when encoded in explicit structural features (Section~\ref{sec:trans_context}). Finally, we present results on ablated models to demonstrate the influence of structural information implicitly encoded in \lstm representations on the final parsing accuracy (Section~\ref{sec:performance_analysis}).

\section{Parsing Model Architecture}
\label{sec:models}

Our graph- and transition-based parsers are based on the \kiparch architecture (see Figure~\ref{fig:architecture}). 
The architecture  has subsequently been extended by, e.g., character-based embeddings \cite{lhoneux-etal:2017:conll} or attention \cite{dozat-manning:2016:arXiv}. To keep the experimental setup clean and simple while focusing on the information flow in the architecture, we abstain from these extensions. We use the basic \kiparch architecture as our starting point with a few minor changes outlined below. For further details we refer the reader to \newcite{kiperwasser-goldberg:2016:TACL-a}.
 
\subsection{Word Representations}

In both transition- and graph-based architectures input tokens are represented in the same way (see level~[1] in Figure~\ref{fig:architecture}). For a given sentence with words $[ w_1, \ldots w_n ]$ and part-of-speech (POS) tags $[t_1, \ldots, t_n]$ each word representation $x_i$ is built from concatenating the embeddings of the word and its POS tag:
\[ x_i = e(w_i) \circ e(t_i) \]

\noindent The embeddings are initialized randomly at training
time and trained together with the model. 

The representations $x_i$ encode words in isolation and do not contain information about their context. 
For that reason they are passed to the \lstm feature extractors (level~[2] in Figure~\ref{fig:architecture}) and represented by a \lstm representation \lstmvec{x_i}:

\[ \lstmvecM{x_i} = \text{\lstm}(x_{1:n}, i) \]

\subsection{Transition-Based Parser}
Transition-based parsers gradually build a tree by applying a sequence of transitions. During training they learn a scoring function for transitions. While decoding they search for the best action given the current state and the parsing history.

Figure~\ref{fig:arch-trans} illustrates the architecture of the transition-based \kiparch parser.
For every configuration $c$ consisting of a stack, buffer, and a set of arcs introduced so far,
the parser 
selects a few core items from the stack and buffer (red arrows in the figure) as features. 
Next, it concatenates their \lstm vectors and passes them to 
a multi-layer perceptron (MLP) which assigns scores to all possible transitions. The highest scoring transition is used to proceed to the next configuration.

Our implementation (denoted \textbf{\tbpars})
uses the arc-standard decoding algorithm \cite{nivre:2004:IncrementalParsing} extended with a \textsc{swap} transition (\aswap, \newcite{nivre:2009:ACLIJCNLP}) to handle non-projective trees. The system applies arc transitions between the two topmost items of the stack (denoted $s_0$ and $s_1$). We use the lazy \textsc{swap} oracle by \newcite{nivre-etAl:2009:IWPT} for training. Labels are predicted together with the transitions.
We experiment with two models with different feature sets:

\textbf{\asslazy{}:} is the simple architecture which does not use structural features. 
Since \newcite{shi-etal:2017:EMNLP} showed that the feature set $\{\lstmvecM{s_0}, \lstmvecM{s_1}, \lstmvecM{b_0} \}$ is minimal for the arc-standard system (i.e., it suffers almost no loss in performance in comparison to larger feature sets but significantly out-performs a feature set built from only two vectors) we apply the same feature set to \aswap. Later we analyze if the set could be further reduced.

\textbf{\asslazyExt{}:} is the extended architecture. We use the original extended feature set from \kiparch: $\{\lstmvecM{s_0}$, $\lstmvecM{s_1}$, $\lstmvecM{s_2}$, $\lstmvecM{b_0}$, $\lstmvecM{s_{0L}}$, $\lstmvecM{s_{0R}}$, $\lstmvecM{s_{1L}}$, $\lstmvecM{s_{1R}}$, $\lstmvecM{s_{2L}}$, $\lstmvecM{s_{2R}}$, $\lstmvecM{b_{0L}}\}$, where $._{L}$ and $._{R}$ denote left- and right-most child.

\begin{table*}[h]
\footnotesize
\centering
	\begin{tabu}{@{}lllllllllllll@{}}
		\toprule
		\rowfont[c]{} & avg. & en-ptb & ar & en & fi & grc & he & ko & ru & sv & zh \\ \midrule
		\asslazy & 76.43 & 90.25 & 76.22 & \sigfive{81.85} & \sigfive{72.51} & \sigfive{71.92} & \sigfive{79.41} & 64.39 & \sigfive{74.35} & \sigfive{80.11} & \sigfive{73.28} \\
		\asslazyExt & 75.56 & 90.25 & 75.77 & 80.50 & 71.47 & 70.32 & 78.62 & 63.88 & 73.82 & 78.80 & 72.17 \\
		\midrule
		\eisner & 77.74 & 91.40 & 77.25 & 82.53 & 74.37 & 73.48 & 80.83 & 65.47 & 76.43 & 81.22 & 74.47 \\
		\eisnerExt & 77.89 & 91.59 & 77.21 & 82.65 & 74.44 & 73.20 & 81.03 & 65.61 & \sigfive{76.79} & 81.42 & \sigfive{74.95} \\
		\bottomrule
	\end{tabu}
\caption{Average (from six runs) parsing results (LAS) on test sets. \sigfiveMarker{} marks statistical significance (\mbox{p-value}~\textless~0.05). Corresponding standard deviations are provided in Table~\ref{tab:standard_dev} in Appendix~\ref{sec:appendix}.}
\label{tab:test}
\end{table*}

\subsection{Graph-Based Parser}

The \kiparch  graph-based parser follows the structured prediction paradigm: while training it learns a scoring function which scores the correct tree higher than all the other possible ones. While decoding it searches for the highest scoring tree for a given sentence. 
The parser employs an arc-factored approach~\cite{mcdonald-EtAl:2005:ACL}, i.e., it decomposes the score of a tree to the sum of the scores of its arcs. 

Figure~\ref{fig:arch-graph} shows the \kiparch graph-based architecture. At parsing time, every pair of words $\langle x_i, x_j\rangle$ yields a \lstm representation 
$\{ \lstmvecM{x_i}, \lstmvecM{x_j} \}$ (red arrows in the figure) which is passed to MLP to compute the score for an arc $x_i \rightarrow x_j$. To find the highest scoring tree we apply \newcite{eisner:1996:CL}'s algorithm. We denote this architecture \textbf{\eisner}. We note in passing that, although this decoding algorithm is restricted to projective trees, it has the advantage that it can be extended to incorporate non-local features while still maintaining exact search in polynomial time.\footnote{Replacing \newcite{eisner:1996:CL}'s algorithm with the Chu-Liu-Edmonds's decoder  \cite{chu-liu:1965,edmonds:1967}  which can predict non-projective arcs causes significant improvements only for the Ancient Greek treebank (1.02 LAS on test set).}

The above-mentioned simple architecture uses a feature set of two vectors $\{ \lstmvecM{h}, \lstmvecM{d} \}$. We extend it and add information about structural context. Specifically, we incorporate information about siblings \lstmvec{s} (blue arrows in the figure). The model follows the second-order model from \newcite{mcdonald-pereira:2006:EACL} and decomposes the score of the tree into the sum of adjacent edge pair scores. We use the implementation of the second-order decoder from \newcite{zhang-zhao:2015:paclic}.
We denote this architecture \textbf{\eisnerExt}.

\section{Experimental Setup}
\label{sec:experimental}

\paragraph{Data sets and preprocessing.}
We perform experiments on a selection of nine treebanks from Universal Dependencies \cite{nivre-EtAl:2016:LREC} (v2.0): Ancient Greek PROIEL (grc), Arabic (ar), Chinese (zh), English (en), Finnish (fi), Hebrew (he), Korean (ko), Russian (ru) and Swedish (sv). This selection was proposed by \newcite{smith-etAl:2018:EMNLP} 
as a sample of languages varying  in language family, morphological complexity, and frequencies of non-projectivity (we refer to \newcite{smith-etAl:2018:EMNLP} for treebank statistics). 
To these 9, we add the English Penn Treebank (en-ptb) converted to Stanford Dependencies.\footnote{We use version 3.4.1 of the Stanford Parser from\\ \url{http://nlp.stanford.edu/software/lex-parser.shtml}} We use sections 2-21 for training, 24 as development set and 23 as test set.

We use automatically predicted universal POS tags in all the experiments. The tags are assigned using a CRF tagger \cite{muller-etAl:2013:EMNLP}. We annotate the training sets via 5-fold jackknifing.

\paragraph{Evaluation.}
We evaluate the experiments using Labeled Attachment Score (LAS).\footnote{The ratio of tokens with a correct head and label to the total number of tokens in the test data.}
We train models for 30 epochs and select the best model based on development LAS. We follow recommendations from \newcite{reimers-gurevych:2018:arXiv} and report averages and standard deviations from six models trained with different random seeds. 
We test for significance using the Wilcoxon rank-sum test with p-value~$<$~0.05.

Analysis is carried out on the development sets in order not to compromise the test sets. We present the results on the concatenation of all the development sets (one model per language).
While the absolute numbers vary across languages, the general trends are consistent with the concatenation.

\paragraph{Implementation details.}
All the described parsers were implemented with the DyNet library \cite{neubig-etal:2017:arXiv}.\footnote{The code can be found on the first author's website.}
We use the same hyperparameters as \newcite{kiperwasser-goldberg:2016:TACL-a} and summarize them in Table~\ref{tab:hyper} in Appendix~\ref{sec:appendix}.

\section{Structural Features and \lstms}
\label{sec:testresults}

\subsection{Simple vs.  Extended Architectures}
We start by evaluating the performance of our four models. The purpose is to verify that the simple architectures will compensate for the lack of additional structural features and achieve comparable accuracy to the extended ones.

\begin{figure}[tb]
 \centering
  \includegraphics[width=0.49\textwidth]{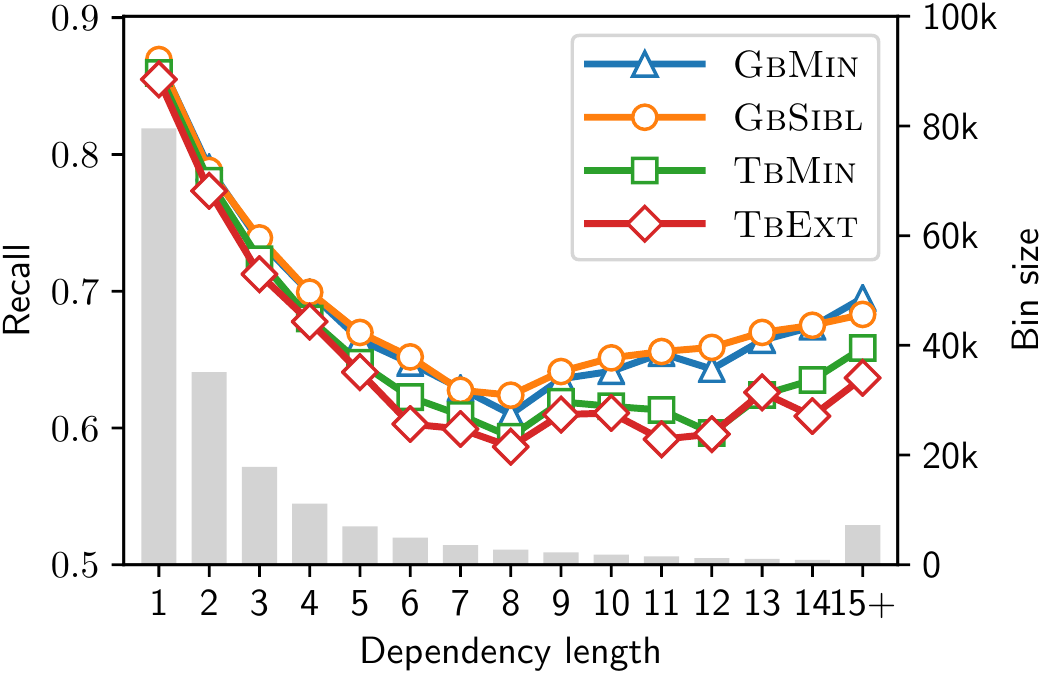}
  \caption{Dependency recall relative to arc length on development sets. The corresponding plot for precision shows similar trends (see Figure~\ref{fig:compare_prec} in Appendix~\ref{sec:appendix}).}
  \label{fig:comp_arcs}
\end{figure}

Table~\ref{tab:test} shows the accuracy of all the parsers. Comparing the simple and extended architectures we see that dropping the structural features does not hurt the performance, neither   for transition-based nor graph-based parsers. Figure~\ref{fig:comp_arcs} displays the 
accuracy relative to dependency length in terms of  recall.\footnote{Dependency recall is defined as the percentage of correct predictions among gold standard arcs of length $l$ \cite{mcdonald-nivre:2007:emnlp}.} It shows that the differences between models are not restricted to arcs of particular lengths.

In the case of graph-based models (\eisner vs. \eisnerExt) adding the second-order features to a \lstm-based parser improves the average performance slightly. However, the difference between those two models is significant only for two out of ten treebanks.
For the transition-based parser (\asslazy vs. \asslazyExt) a different effect can be noticed -- additional features cause a significant loss in accuracy for seven out of ten treebanks. One possible explanation might be that \asslazyExt suffers more from error propagation than \asslazy. The parser is greedy and after making the first mistake it starts drawing features from configurations which were not observed during training.
Since the extended architecture uses more features than the simple one the impact of the error propagation might be stronger. This effect can be noticed in Figure~\ref{fig:comp_arcs}. The curves for \asslazy and \asslazyExt are almost parallel for the short arcs but the performance of \asslazyExt deteriorates for the longer ones, which are more prone to error propagation \cite{mcdonald-nivre:2007:emnlp}.

\subsection{Influence of \lstms}
\label{sec:feat_and_lstms}

We now investigate whether \lstms are the reason for models being able to compensate for lack of features drawn from partial subtrees.

\begin{figure*}[t]
\centering
\begin{subfigure}[b]{.49\textwidth}
  \centering
  \includegraphics[width=0.98\linewidth]{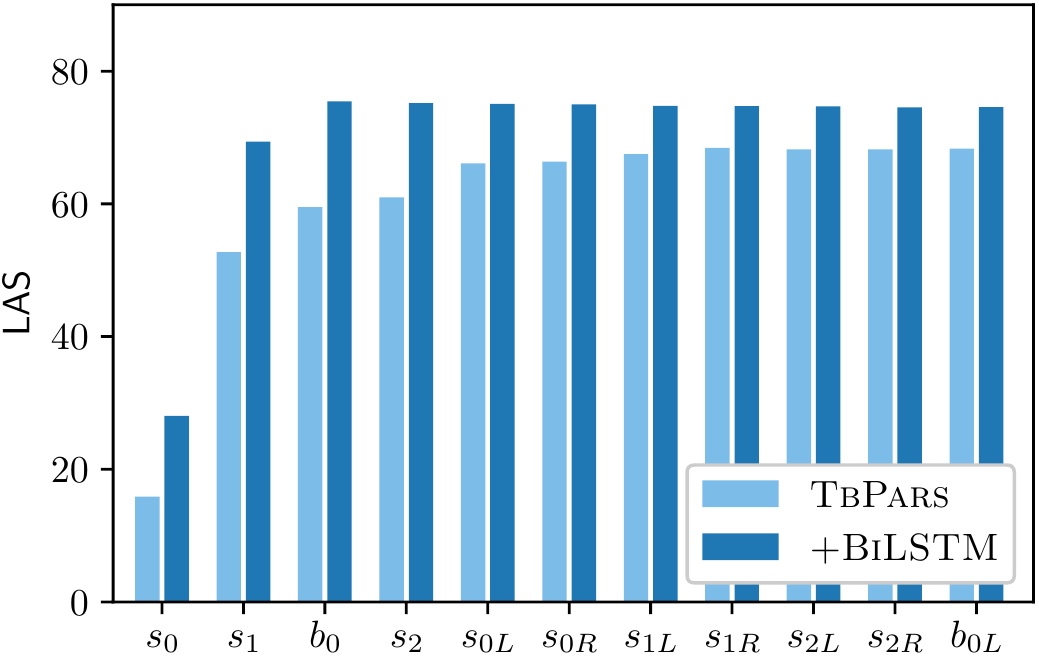}
  \caption{Transition-based parser}
  \label{fig:trans_feat}
\end{subfigure}\hfill
\begin{subfigure}[b]{.485\textwidth}
  \centering
  \includegraphics[width=0.98\linewidth]{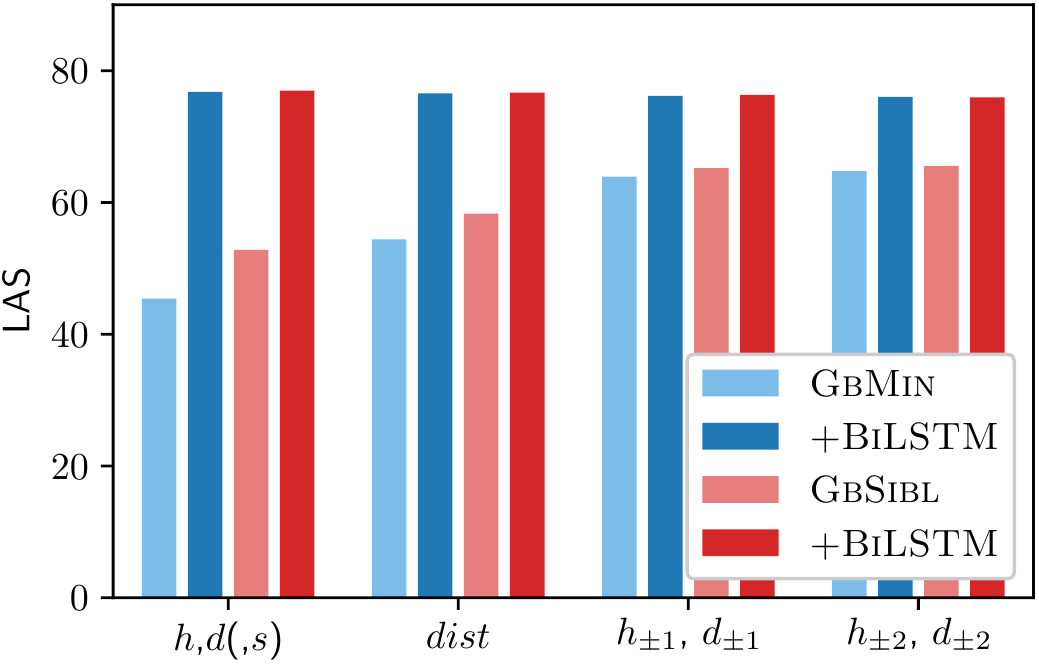}
  \caption{Graph-based parser}
  \label{fig:graph_las}
\end{subfigure}
\caption{Parsing accuracy (average LAS over ten treebanks) with incremental extensions to the feature set. }
\label{fig:surf_impact}
\end{figure*}

\paragraph{Transition-based parser.}
We train \tbpars in two settings: with and without \lstms (when no \lstms are used we pass vectors $x_i$ directly to the MLP layer following \newcite{chen-manning:2014:EMNLP})
and with different feature sets. 
We start with a feature set $\{ s_0 \}$ and consecutively add more until we reach the full feature model of \asslazyExt{}.

%% without bi-lstms

Figure~\ref{fig:trans_feat} displays the accuracy of all the trained models. First of all, we notice that the models without \lstms (light bars) benefit from structural features. The biggest gains in the average performance are visible after adding vectors $s_{0L}$ (5.15 LAS) and $s_{1R}$ (1.12 LAS) . After adding  $s_{1R}$ the average improvements become modest.

%% after adding bi-lstms

Adding the \lstm representations changes the picture (dark bars). First of all, as in the case of arc-standard system \cite{shi-etal:2017:EMNLP}, the feature set $\{ \lstmvecM{s_0}, \lstmvecM{s_1}, \lstmvecM{b_0}\}$ is minimal for \aswap: none of the other structural features are able to improve the performance of the parser but dropping $b_0$ causes a big drop of almost 6 LAS on average.  Secondly, the parsers which use \lstms always have a big advantage over the parsers which do not, regardless of the feature model used.

%% graph parsers

\paragraph{Graph-based parser.}
We train two models: \eisner and \eisnerExt with and without \lstms. 
To ensure a fairer comparison with the models without \lstms we expand the basic feature sets ($\{ \lstmvecM{h}, \lstmvecM{d}\}$ and 
$\{ \lstmvecM{h}, \lstmvecM{d}, \lstmvecM{s}\}$) 
with additional surface features known from classic graph-based parsers, such as distance between head and dependent ($dist$), words at distance of~1 from heads and dependents ($h_{\pm1}$, $d_{\pm1}$) and at distance~$\pm2$. We follow \newcite{wang-chang:2016:ACL} and encode distance as randomly initialized embeddings.

%% without bi-lstms

Figure~\ref{fig:graph_las} displays the accuracy of all the trained models with incremental extensions to their feature sets. First of all, we see that surface features ($dist$, $h_{\pm1}$, $d_{\pm1}$, $h_{\pm2}$, $d_{\pm2}$) are beneficial for the models without \lstm representations (light bars). The improvements are visible for both parsers, with the smallest gains after adding $h_{\pm2}$, $d_{\pm2}$ vectors: on average 0.35 LAS for \eisnerExt and 0.83 LAS for \eisner.

%% adding bi-lstms

As expected, adding \lstms changes the picture. 
Since the representations capture surface context, they already contain a lot of information about words around heads and dependents and adding features $h_{\pm1}$, $d_{\pm1}$ and $h_{\pm2}$, $d_{\pm2}$ does not influence the performance. Interestingly, 
introducing $dist$ is also redundant which suggests that either \lstms are aware of the distance between tokens or they are not able to use this information in a meaningful way. 
%do not know how to use it even if introduced explicitly.
Finally, even after adding all the surface features the models which do not employ \lstms are considerably behind the ones which do.

%% adding structure

Comparing \eisner (blue) with \eisnerExt (red) we see that adding information about structural context through second-order features is beneficial  when the \lstm are not used (light bars): 
the second-order \eisnerExt has an advantage over \eisner of 0.81 LAS even when both of the models use all the additional surface information (last group of bars on the plot).   
%the latter, second-order model which encodes more information about the structural context is suitable for 
%the \kiparch architecture. 
But this advantage drops down to insignificant 0.07 LAS when the \lstms are incorporated.

\paragraph{}
We conclude that, for both transition- and graph-based parsers, \lstms not only compensate for absence of structural features but they also encode more information than provided by the manually designed feature sets.

\section{Implicit Structural Context}

Now that we have established that structural features are indeed redundant for models which employ \lstms we examine the ways in which the simple parsing models (\asslazy and \eisner) implicitly encode information about partial subtrees.

\subsection{Structure and \lstm Representations}
\label{sec:representation_analysis}

We start by looking at the \lstm representations. We know that the representations are capable of capturing syntactic relations when they are trained on a syntactically related task, e.g, number prediction task \cite{linzen-etal:2016:TACL}. We evaluate how complicated those relations can be when the representations are trained together with a dependency parser.

\begin{figure*}[t]
\centering
\begin{subfigure}[b]{.49\textwidth}
  \centering
  \includegraphics[width=0.98\linewidth]{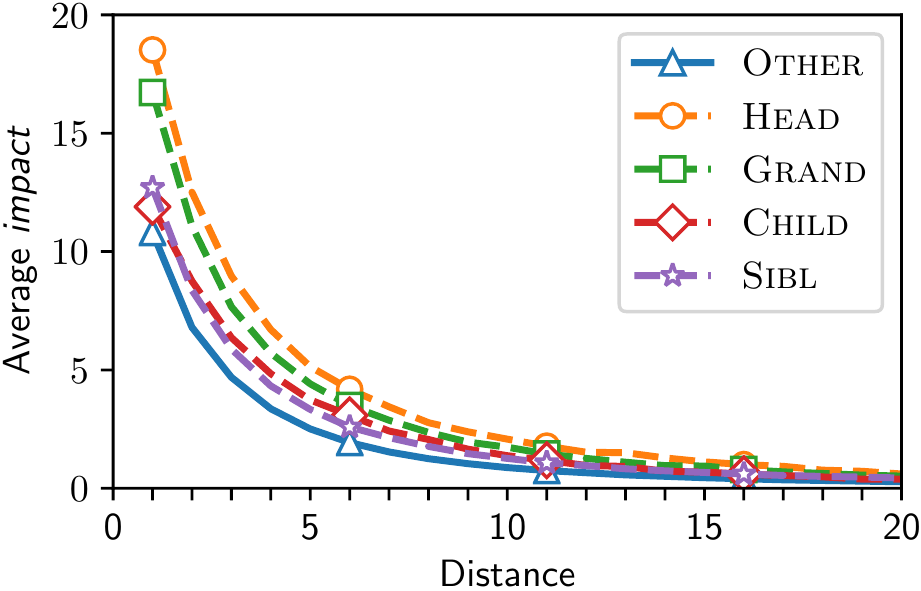}
  \caption{Transition-based parser (\asslazy)}
  \label{fig:trans_surf}
\end{subfigure}\hfill
\begin{subfigure}[b]{.49\textwidth}
  \centering
  \includegraphics[width=0.98\linewidth]{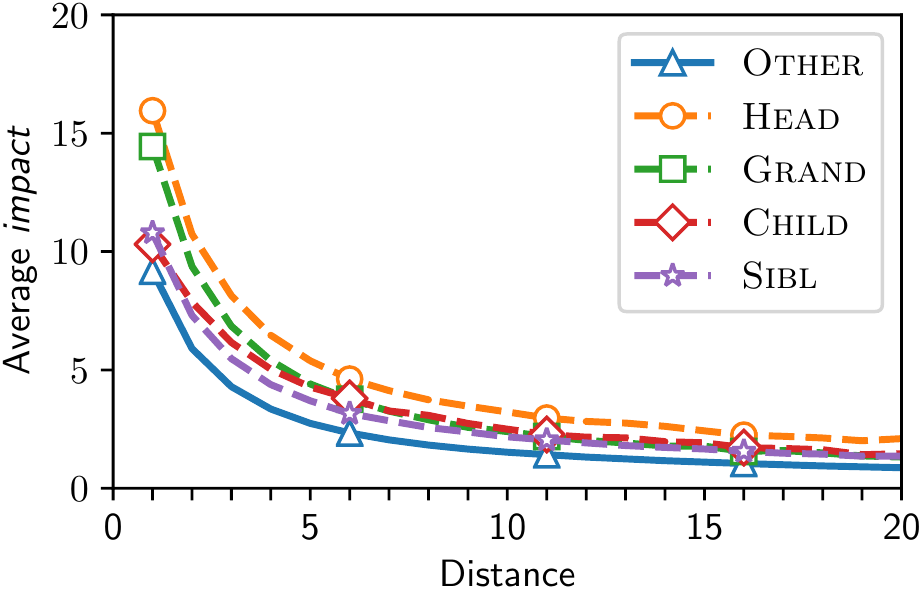}
  \caption{Graph-based parser (\eisner)}
  \label{fig:graph_surf}
\end{subfigure}
\caption{The average impact of tokens on \lstm vectors trained with dependency parser with respect to the surface distance and the structural (gold-standard) relation between them.
}
\label{fig:impact_feats}
\end{figure*}

To do so, we follow \newcite{gaddy-etAl:2018:NAACL} and use derivatives to estimate how sensitive a particular part of the architecture is with respect to changes in input. Specifically, for every vector $\lstmvecM{x}$ we measure how it is influenced by every word representation $x_i$ from the sentence. If the derivative of $\lstmvecM{x}$ with respect to $x_i$ is high then the word $x_i$ has a high influence on the vector. We compute the $l_2$-norm of the gradient of $\lstmvecM{x}$ with respect to $x_i$ and normalize it by the sum of norms of all the words from the sentence calling this measure \textbf{\textit{\impactT}}:

\[ \impactLSTM(\lstmvecM{x}, i) = 100 \times \frac{ || \frac{\partial \lstmvecM{x}}{\partial x_i} || } { \sum_j || \frac{\partial \lstmvecM{x}}{\partial x_j} || } \]

For every sentence from the development set and every vector \lstmvec{x_i} we calculate the impact of every representation ${x_j}$  from the sentence on the vector \lstmvec{x_i}. We bucket those impact values according to the distance between the representation and the word. We then use the gold-standard trees to divide every bucket into five groups: correct heads of $x_i$, children (i.e., dependents) of $x_i$, grandparents (i.e., heads of heads), siblings, and other. 

Figure~\ref{fig:impact_feats} shows the average impact of tokens at particular positions. 
Similarly as shown by \newcite{gaddy-etAl:2018:NAACL} even words 15 and more positions away have a non-zero effect on the \lstm vector.
Interestingly, the impact of words which we know to be structurally close to $x_i$ is higher. For example, for the transition-based parser (Figure~\ref{fig:trans_surf}) at positions $\pm5$ an average impact is lower than 2.5\%, children and siblings of $x_i$ have a slightly higher impact, and the heads and grandparents around 5\%. 
For the graph-based parser (Figure~\ref{fig:graph_surf}) the picture is similar with two noticeable differences. 
The impact of heads is much stronger for words 10 and more positions apart. But it is smaller than in the case of transition-based parser when the heads are next to $x_i$. % (compare positions $\pm1$  or $\pm2$ at Figures . for example, at positions

\paragraph{}
We conclude that the \lstms  are indeed 
influenced by the distance, but when trained with a dependency parser they also capture 
a significant amount of
non-trivial syntactic relations.

\subsection{Structure and Information Flow}
\label{sec:trans_context}

\begin{figure*}[t]
\centering
\begin{subfigure}[b]{.582\textwidth}
  \centering
  \includegraphics[width=0.95\linewidth]{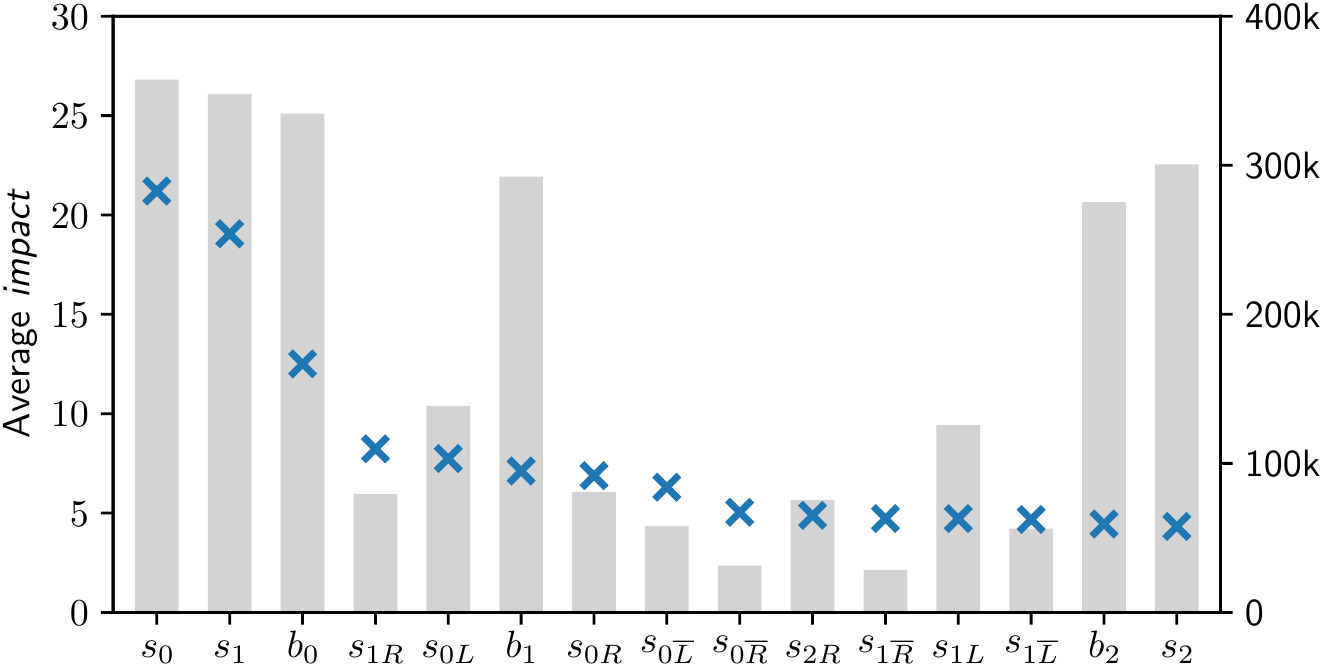}
  \caption{Transition-based parser  (\asslazy); positions depend on the \\ configuration; $._{\overline{L}}$ marks left children that are not the leftmost, \\ $._{\overline{R}}$ marks right children that are not the rightmost. \\ \vphantom{bla}}
  \label{fig:trans_impact}
\end{subfigure}\hfill
\begin{subfigure}[b]{.41\textwidth}
  \centering
  \includegraphics[width=0.95\linewidth]{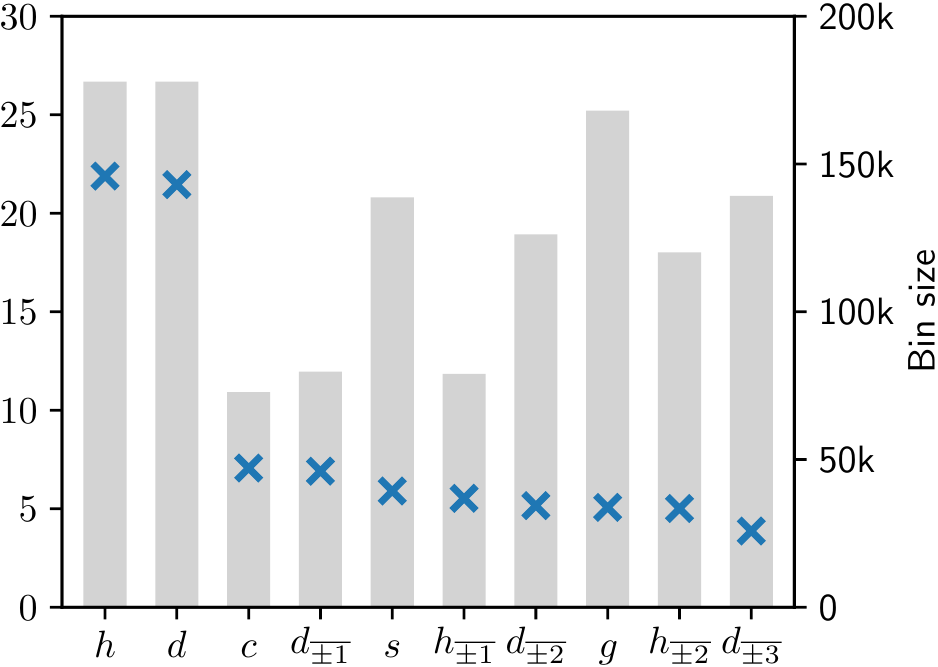}
  \caption{Graph-based parser  (\eisner); positions are: heads (h), dependents (d), children of d (c), siblings (s), grandparents~(g), h,d$_{\overline{\pm i}}$ tokens at distance $\pm i$ from h or d which are none of h, d, c, s, or g. }
  \label{fig:graph_impact}
\end{subfigure}
\caption{Positions with the highest impact on the MLP scores (blue crosses) and their frequency (gray bars).}
\label{fig:context_impact}
\end{figure*}

Now that we know that the representations encode structural information we ask how this information influences the decisions of the parser. 

First, we investigate how much structural information flows into the final layer of the network. When we look back at the architecture in Figure~\ref{fig:architecture} we see that when the final MLP scores possible transitions or arcs it uses only feature vectors $\{ \lstmvecM{s_0}, \lstmvecM{s_1}, \lstmvecM{b_0} \}$ or $\{ \lstmvecM{h}, \lstmvecM{d}\}$. But thanks to the \lstms  the vectors encode information about other words from the sentence. We examine from which words the signal is the strongest when the parser makes the final decision.

We extend the definition of \textit{\impactLSTM} to capture how a specific word representation $x_i$ influences the final MLP score $sc$ (we calculate the derivative of $sc$ with respect to $x_i$). We parse every development sentence. 
% from the  set. 
For every predicted transition/arc we calculate how much its score $sc$ was affected by every word from the sentence. We group impacts of words depending on their positions. % in the sentence.

\paragraph{Transition-based parser.}
For the transition-based parser we group tokens according to their positions in the configuration. For example, for the decision in Figure~\ref{fig:arch-trans} $\impactT(sc, 1)$ would be grouped as $s_1$ and $\impactT(sc, j)$ as $s_{0R}$. 

In Figure~\ref{fig:trans_impact} we plot the 15 positions with the highest impact and the number of configurations they appear in (gray bars). As expected, $s_0$, $s_1$, and $b_0$ have the highest influence on the decision of the parser. The next two positions are $s_{1R}$ and $s_{0L}$. Interestingly, those are the same positions which used as features caused the biggest gains in 
performance for the models which \textit{did not} use \lstms (see Figure~\ref{fig:trans_feat}). They are much less frequent than $b_1$ but when they are present the model is strongly influenced by them. 
After $b_1$ we can notice positions which are not part of the manually designed extended feature set of \asslazyExt, such 
as $s_{0\overline{L}}$ (left children of $s_0$ that are not the leftmost).

\paragraph{Graph-based parser.}
For the graph-based parser we group tokens according to their position in the full predicted tree.
We then bucket the impacts into: heads (h), dependents (d), children (i.e., dependents of dependents)~(c), siblings~(s), and grandparents (i.e., heads of heads) (g). 
Words which do not fall into any of those categories are grouped according to their surface distance from heads and dependents. For example, \hPMTwo are tokens two positions away from the head which do not act as dependent, child, sibling, or grandparent.

Figure~\ref{fig:graph_impact} presents 10 positions with the highest impact and the number of arcs for which they are present (gray bars). As expected, heads and dependents have the highest impact on the scores of arcs, much higher than any of the other tokens. Interestingly, among the next three bins with the highest impact are children and siblings. Children are less frequent than structurally unrelated tokens at distance 1  (\hPMOne, \dPMOne), and much less frequent than \hPMTwo or \dPMTwo but they influence the final scores more. The interesting case is siblings -- they not only have a strong average impact but they are also very frequent, suggesting that they are very important for the parsing accuracy.

\paragraph{}
The results above show that the implicit structural context is not only present in the models, but also more diverse than when incorporated through conventional  explicit structural features.
 
\subsection{Structure and Performance}
\label{sec:performance_analysis}

\begin{figure*}[t]
\centering
\begin{subfigure}[b]{.54\textwidth}
  \centering
  \includegraphics[width=0.95\linewidth]{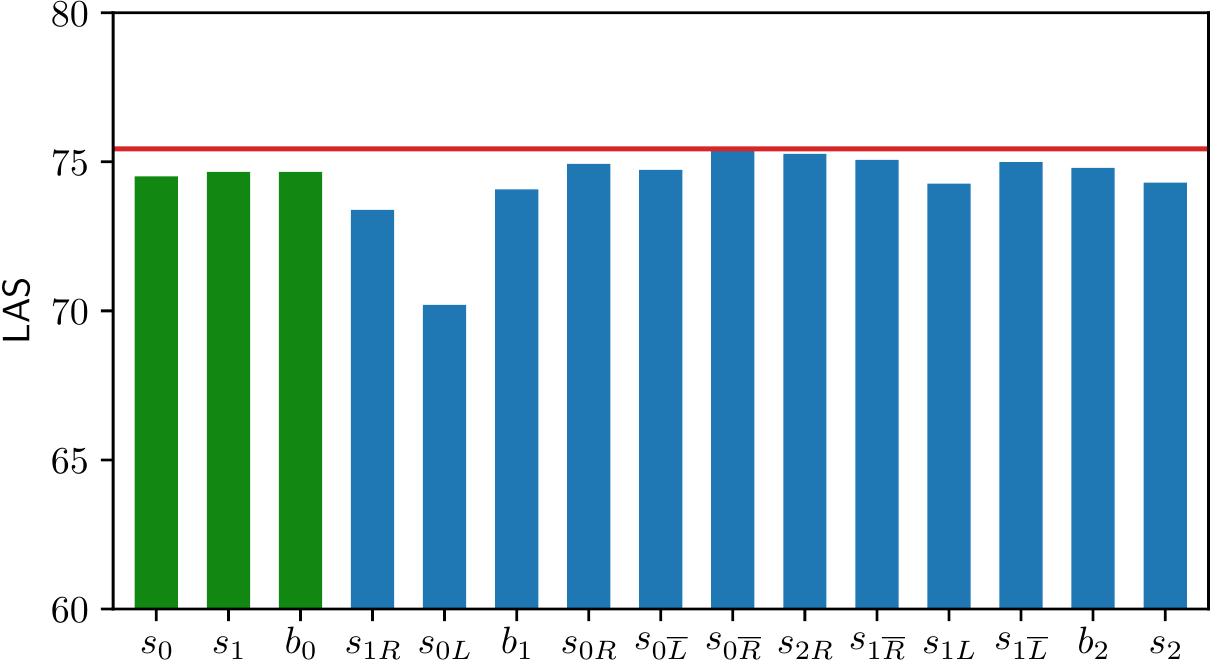}
  \caption{Transition-based parser  (\asslazy)}
  \label{fig:trans_drop}
\end{subfigure}\hfill
\begin{subfigure}[b]{.45\textwidth}
  \centering
  \includegraphics[width=0.79\linewidth]{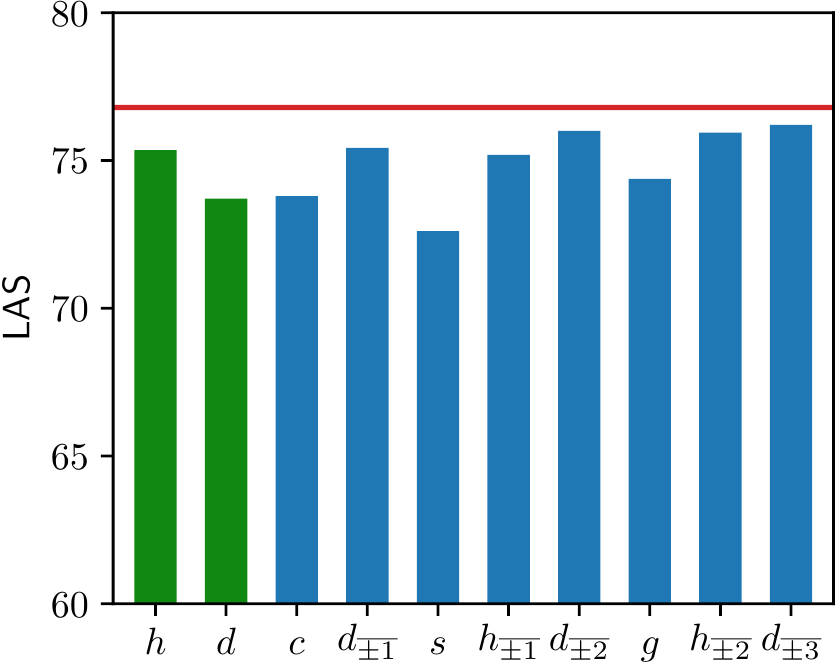}
  \caption{Graph-based parser  (\eisner)}
  \label{fig:graph_drops}
\end{subfigure}
\caption{The performance drops when tokens at particular positions are removed from the \lstm encoding. The red line marks average LAS of uninterrupted model. 
Feature sets of both models are highlighted in green. }
\label{fig:perf_drops}
\end{figure*}

Finally, we investigate if the implicit structural context is important for the performance of the parsers. To do so, we take tokens at structural positions with the highest impact and train new ablated models in which the information about those tokens is dropped from the \lstm layer. For example, while training an ablated model without $s_{0L}$, for every configuration we re-calculate all the \lstm vectors as if $s_{0L}$ was not in the sentence. When there is more than one token at a specific position, for example \sLRest{0} or $c$ (i.e., children of the dependent), we pick a random one to drop. That way every ablated model looses information about at most one word.

We note that several factors can be responsible for drops in performance of the ablated models. For example, the proposed augmentation distorts distance between tokens which might have an adverse impact on the trained representations. Therefore, in the following comparative analysis we interpret 
the obtained drops as an approximation of how much
%under comparable conditions, 
particular tokens influence the performance of the models. 

%
%\footnote{Several factors can be responsible for drops in performance of the ablated models. For example, the proposed ablation requires multiple re-calculations of the representations which might have a negative effect on training. However, the main purpose of  those experiments is to set side by side different structural positions and compare their influence on the parsing performance. The employed augmentation allows us to perform this comparison under comparable conditions for all structural positions.}

\paragraph{Transition-based parser.}
Figure~\ref{fig:trans_drop} presents the drops in the parsing performance for the ablated models.\footnote{It is worth noting that not all of the models suffer from the ablation. For example, dropping vectors $s_{2R}$ causes almost no harm. This suggests that re-calculating the representations  multiple times does not have a strong negative effect on training.}
First of all, removing the vectors $\{ \lstmvecM{s_0}, \lstmvecM{s_1}, \lstmvecM{b_0} \}$ (marked in green on the plot) only from the \lstm layer (although they are still used as features) causes visible drops in performance. 
One explanation might be that when the vector $\lstmvecM{s_0}$ is recalculated without knowledge of $s_1$ the model loses information about the distance between them.
Secondly, we can notice that other drops depend on both the impact and frequency of positions. The biggest declines are visible after removing $s_{0L}$ and  $s_{1R}$ -- precisely the positions which we found to have the highest impact on the parsing decisions.
Interestingly, the positions which were not a part of the \asslazyExt feature set, such as \sLRest{0} or \sRRest{1}, although not frequent are important for the performance.

\paragraph{Graph-based parser.}
Corresponding results for the graph-based parser 
are presented in Figure~\ref{fig:graph_drops} 
(we use gold-standard trees as the source of information about structural relations between tokens).
The biggest drop 
can be observed for ablated models without siblings. Clearly, information coming from those tokens implicitly into MLP is very important for the final parsing accuracy.
The next two biggest drops are caused by lack of children and grandparents.
As we showed in Figure~\ref{fig:graph_impact} children, although less frequent, have a stronger impact on the decision of the parser. But dropping 
grandparents also significantly harms the models.

\paragraph{}
We conclude that information about partial subtrees is not only present when the parser makes final decisions but also strongly influences those decisions.
Additionally, the deteriorated accuracy of the ablated models shows that the implicit structural context can not be easily compensated for.

\section{Related Work}
\label{sec:background}

\paragraph{Feature  extraction.}
\newcite{kiperwasser-goldberg:2016:TACL-a} and \newcite{cross-huang:2016:ACL} first applied \lstms to extract features for transition-based dependency parsers. The authors demonstrated that an architecture using only a few positional features (four for the arc-hybrid system and three for arc-standard) is sufficient to achieve state-of-the-art performance.
\newcite{shi-etal:2017:EMNLP} showed that this number can be further reduced to two features for arc-hybrid and arc-eager systems.
Decreasing the size of the feature set not only allows for construction of lighter and faster neural networks \cite{wang-chang:2016:ACL,vilares-gomez-rodriguez:2018:udw} but also enables the use of exact search algorithms for several projective \cite{shi-etal:2017:EMNLP} and 
non-projective \cite{gomez-rodriguez-etal-2018-global}  transition systems.
A similar trend can be observed for graph-based dependency parsers. State-of-the-art models \cite{kiperwasser-goldberg:2016:TACL-a, dozat-manning:2016:arXiv} typically use only two features of heads and dependents, possibly also incorporating their distance \cite{wang-chang:2016:ACL}. Moreover, \newcite{wang-chang:2016:ACL} show that arc-factored \lstm-based parsers can compete with conventional higher-order models in terms of accuracy.
 
None of the above mentioned efforts address the question how dependency parsers are able to compensate for the lack of structural features. 
The very recent work by \newcite{lhoneux-etal:2019:arXiv} looked into this issue from a different perspective than ours -- composition. They showed that composing the structural context with recursive networks as in \newcite{dyer-etal:2015:acl} is redundant for the \kiparch transition-based architecture. The authors analyze components of the \lstms to show which of them (forward v. backward LSTM) is responsible for capturing subtree information.

\paragraph{RNNs and syntax.}
Recurrent neural networks, which \lstms are a variant of, have been repeatedly analyzed to understand whether they can learn syntactic relations. Such analyses differ in terms of: (1) methodology they employ to probe what type of knowledge the representations learned and (2) tasks on which the representations are trained on.
\newcite{shi-etal:2016:EMNLP} demonstrated that sequence-to-sequence machine-translation systems capture source-language syntactic relations. \newcite{linzen-etal:2016:TACL} showed that when trained on the task of number agreement prediction the representations capture a non-trivial amount of grammatical structure (although recursive neural networks  are better at this task than  sequential LSTMs \cite{kuncoro-etal:2018:ACL}).
\newcite{blevins-etal-2018-deep} found that RNN representations trained on a variety of NLP tasks (including dependency parsing) are able to induce syntactic features (e.g., constituency labels of parent or grandparent) even without explicit supervision.
Finally, \newcite{conneau-etal:2018:ACL} designed a set of tasks probing linguistic knowledge of sentence embedding methods.

Our work contributes to this line of research in two ways: (1) from the angle of methodology, we show how to employ derivatives to pinpoint what syntactic relations the representations learn; (2) from the perspective of tasks, we demonstrate how \lstm-based dependency parsers take advantage of structural information encoded in the representations.
In the case of constituency parsing \newcite{gaddy-etAl:2018:NAACL} offer such an analysis. The authors show that their \lstm-based models implicitly learn the same information which was conventionally provided to non-neural parsers, such as grammars and lexicons. 

\section{Discussion and Conclusion}
\label{sec:conclusion}

We examined how the application of \lstms influences the modern transition- and graph-based parsing architectures. The \lstm-based
parsers can compensate for the lack of traditional structural features. Specifically, the features drawn from partial subtrees become redundant because the parsing models encode them implicitly.

The main advantage of \lstms comes with their ability to capture not only surface but also syntactic relations. When the representations are trained together with a parser they encode 
structurally-advanced relations such as heads, children, or even siblings and grandparents. This structural information is then passed directly (through feature vectors) and indirectly (through \lstms encoding) to MLP and is used for scoring transitions and arcs. 
Finally, the implicit structural information is important for the final parsing decisions: dropping it in ablated models causes their performance to deteriorate.

The introduction of \lstms into dependency parsers has an additional interesting consequence. 
The classical transition- and graph-based dependency parsers have their strengths and limitations due to the trade-off between the richness of feature functions and the inference algorithm \cite{mcdonald-nivre:2007:emnlp}.
Our transition- and graph-based architectures use the same word representations. We showed that those representations trained together with the parsers capture syntactic relations in a similar way. Moreover, the transition-based parser does not incorporate structural features through the feature set. And the graph-based parser makes use of far away surface tokens but also structurally related words.
Evidently, the employment of \lstm feature extractors blurs the difference between the two architectures. The one clear advantage of the graph-based parser is that it performs global inference
(but exact search algorithms are already being applied to projective \cite{shi-etal:2017:EMNLP} and non-projective \cite{gomez-rodriguez-etal-2018-global} transition systems). 
Therefore, an interesting question is if integrating those two architectures can still be beneficial for the parsing accuracy as in \newcite{nivre-mcdonald:2008:ACL}.
We leave this question for future work.

\section*{Acknowledgments}

This work was supported by the Deutsche Forschungsgemeinschaft (DFG) via the SFB 732, project D8.
We would like to thank the anonymous reviewers for their comments. We also
thank our colleagues Anders Bj\"{o}rkelund, \"{O}zlem \c{C}etino\u{g}lu, and  Xiang Yu for many conversations and comments on this work.

\bibliography{dnpaper}
\bibliographystyle{./acl2019-latex/acl_natbib}

\appendix

\onecolumn
\section{Appendix}
\label{sec:appendix}

\begin{table}[htb]
\footnotesize
\centering
	\begin{tabu}{ll}
		\toprule
		Word embedding dimension  & 100 \\
		POS tag embedding dimension & 20 \\
		Hidden units in MLP & 100 \\
		\lstm layers & 2 \\
		\lstm dimensions &  125 \\
		$\alpha$ for word dropout & 0.25 \\
		Trainer & Adam \\
		Non-lin function & tanh \\
		\bottomrule
	\end{tabu}
\caption{Hyperparameters for the parsers.}
\label{tab:hyper}
\end{table}

\begin{table}[h]
\footnotesize
\centering
	\begin{tabu}{@{}llllllllllll@{}}
		\toprule
		\rowfont[c]{} & en-ptb & ar & en & fi & grc & he & ko & ru & sv & zh \\ \midrule
		\asslazy & 0.237 & 0.323 & 0.207 & 0.163 & 0.382 & 0.391 & 0.740 & 0.282 & 0.295 & 0.398 \\
		\asslazyExt & 0.211 & 0.191 & 0.176 & 0.323 & 0.472 & 0.454 & 0.456 & 0.408 & 0.257 & 0.267 \\
		\midrule
		
		\eisner & 0.146 & 0.179 & 0.212 & 0.157 & 0.340 & 0.269 & 0.300 & 0.228 & 0.379 & 0.408 \\
		\eisnerExt & 0.103 & 0.186 & 0.149 & 0.219 & 0.372 & 0.229 & 0.163 & 0.169 & 0.195 & 0.441 \\
		\bottomrule
	\end{tabu}
\caption{Standard deviation for results in Table~\ref{tab:test}.}
\label{tab:standard_dev}
\end{table}

\begin{figure}[h]
  \centering
  \includegraphics[width=0.5\linewidth]{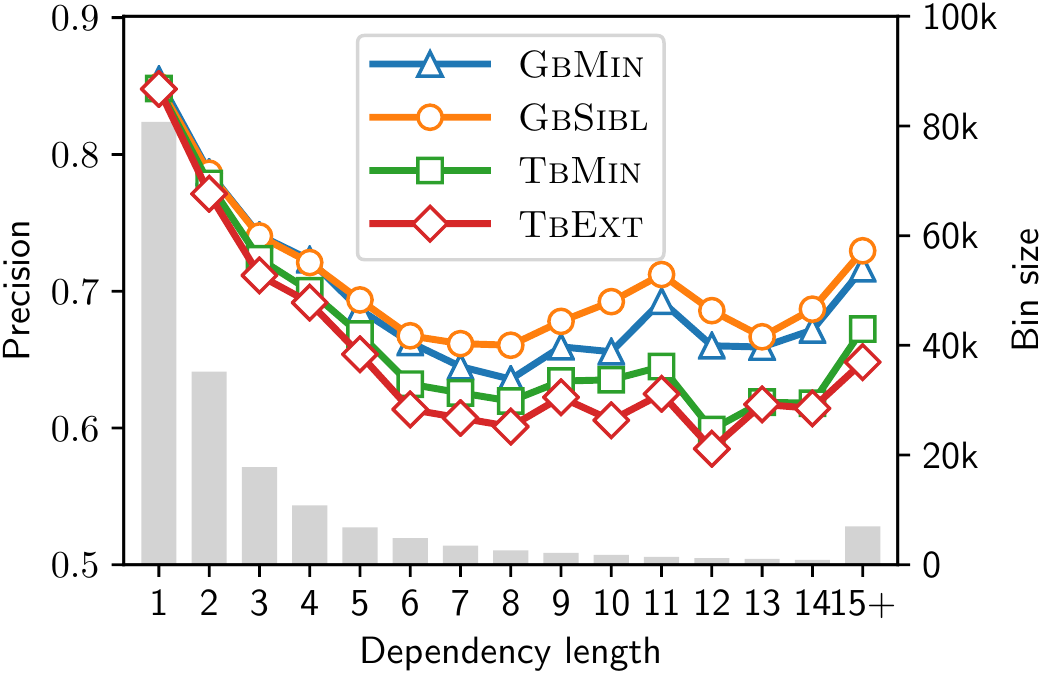}
  \captionof{figure}{Dependency precision relative to arc length on development sets.}
  \label{fig:compare_prec}
\end{figure}

\end{document}